\documentclass[sigconf]{acmart}

\usepackage{epsfig}
\usepackage{graphicx}
\usepackage{amsmath}
\usepackage{helvet}  %Required
\usepackage{courier}  %Required
\usepackage{url}  %Required
\usepackage{xcolor}
\usepackage{multirow}
\usepackage{subcaption}
\usepackage{array}

\usepackage{color, colortbl}
\definecolor{mygray}{gray}{0.9}

\newcommand{\blfootnote}[1]{%
  \begingroup
  \renewcommand\thefootnote{}\footnote{#1}%
  \addtocounter{footnote}{-1}%
  \endgroup
}

\newcolumntype{P}[1]{>{\centering\arraybackslash}p{#1}}

\def\BibTeX{{\rm B\kern-.05em{\sc i\kern-.025em b}\kern-.08emT\kern-.1667em\lower.7ex\hbox{E}\kern-.125emX}}
    
\begin{document}
\acmSubmissionID{3}
%
% The "title" command has an optional parameter, allowing the author to define a "short title" to be used in page headers.
% \title{A Novel Multimedia-based method for Assisting Ophthalmologists to Diagnose Retinal Diseases}
% \title{A Novel Multimedia-based method for Ophthalmologists to Diagnose Retinal Diseases}
% \title{Query-based Video Summarization with Self-supervision and Contextualized Word Representations}
% \title{Query-based Video Summarization with Self-supervision}
% \title{Improving Operational Efficiency in Law and Finance: Leveraging Large Language Models for Criminal Sentencing and Legacy Estimation}
% \title{Enhancing Legal and Financial Decision-Making and Operational Efficiency through Large Language Models}
% \title{Enhancing Legal Decision-Making and Operational Efficiency through Large Language Models}
% \title{Enhancing Decision-Making and Operational Efficiency in Legal Domain through Large Language Models}
% \title{Enhancing Decision-Making and Operational Efficiency in the Legal Domain through Large Language Models for Precise Numerical Estimation} 
% \title{Enhancing Numerical Estimation and Operational Efficiency in the Legal Domain through Large Language Models}
% \title{Optimizing Numerical Estimation and Operational Efficiency in the Legal Domain through Large Language Models}
\title[Optimizing Numerical Estimation and Operational Efficiency in the Legal Domain]{Optimizing Numerical Estimation and Operational Efficiency in the Legal Domain through Large Language Models}
% \title{Optimizing Numerical Estimation and Decision-Making in the Legal Domain through Large Language Models}
% \title{Enhancing Numerical Estimation and Decision-Making in the Legal Domain through Large Language Models}
% \title{Enhancing Legal Decision-Making and Numerical Estimation through Large Language Models}
% \title{Enhancing Decision-Making and Operational Efficiency in the Legal Domain through Large Language Models for Precision-Oriented Legal Tasks}
% \title{GPT2MVS: Generative Pretrained Transformer 2 with hierarchical attention for Multi-modal Video Summarization}

\author{Jia-Hong Huang$^*$}
\affiliation{%
  \institution{University of Amsterdam, Netherlands}}
\email{j.huang@uva.nl}

\author{Chao-Chun Yang$^*$}
\affiliation{%
  \institution{Yang Chao-Chun Law Firm, \\ Taiwan}}
\email{lawyerycj@gmail.com}

\author{Yixian Shen}
\affiliation{%
  \institution{University of Amsterdam, Netherlands}}
\email{y.shen@uva.nl}

\author{Alessio M. Pacces}
\affiliation{%
  \institution{University of Amsterdam, Netherlands}}
\email{a.m.pacces@uva.nl}

\author{Evangelos Kanoulas}
\affiliation{%
  \institution{University of Amsterdam, Netherlands}}
\email{E.Kanoulas@uva.nl}

%
% The abstract is a short summary of the work to be presented in the article.
\begin{abstract}
The legal landscape encompasses a wide array of lawsuit types, presenting lawyers with challenges in delivering timely and accurate information to clients, particularly concerning critical aspects like potential imprisonment duration or financial repercussions. Compounded by the scarcity of legal experts, there's an urgent need to enhance the efficiency of traditional legal workflows. Recent advances in deep learning, especially Large Language Models (LLMs), offer promising solutions to this challenge. Leveraging LLMs' mathematical reasoning capabilities, we propose a novel approach integrating LLM-based methodologies with specially designed prompts to address precision requirements in legal Artificial Intelligence (LegalAI) applications. The proposed work seeks to bridge the gap between traditional legal practices and modern technological advancements, paving the way for a more accessible, efficient, and equitable legal system. To validate this method, we introduce a curated dataset tailored to precision-oriented LegalAI tasks, serving as a benchmark for evaluating LLM-based approaches. Extensive experimentation confirms the efficacy of our methodology in generating accurate numerical estimates within the legal domain, emphasizing the role of LLMs in streamlining legal processes and meeting the evolving demands of LegalAI. 
% To foster innovation in LegalAI, the code and dataset will be released publicly.
\end{abstract}


\acmConference[CIKM '24]{Proceedings of the 33rd ACM International Conference on Information and Knowledge Management}{Oct. 21--25}{USA}
%
% Keywords. The author(s) should pick words that accurately describe the work being
% presented. Separate the keywords with commas.
%
% A "teaser" image appears between the author and affiliation information and the body 
% of the document, and typically spans the page. 

%   \includegraphics[width=\textwidth]{model_flowchart-AAAI_new.pdf}
%   \caption{
%     (a) is an existing traditional medical treatment procedure for retinal diseases. Please refer to the INTRODUCTION section for more details. In (b), we propose this multimedia-based method to improve (a) based on the point-of-care (POC) \cite{pai2012point} concept. In the proposed method, it mainly contains three modules, including DNN-based, DNN visual explanation, and multimedia visualization modules. Please refer to the METHOD section for the detail explanation. DNN indicates Deep Neural Networks.
%     }
% %   \Description{Enjoying the baseball game from the third-base seats. Ichiro Suzuki preparing to bat.}
%   \label{fig:figure1}

%
% This command processes the author and affiliation and title information and builds
% the first part of the formatted document.
\maketitle
\blfootnote{$*$ Equal contribution.}

\section{Introduction}

\begin{figure*}[t!]
\begin{center}
\includegraphics[width=1.0\linewidth]{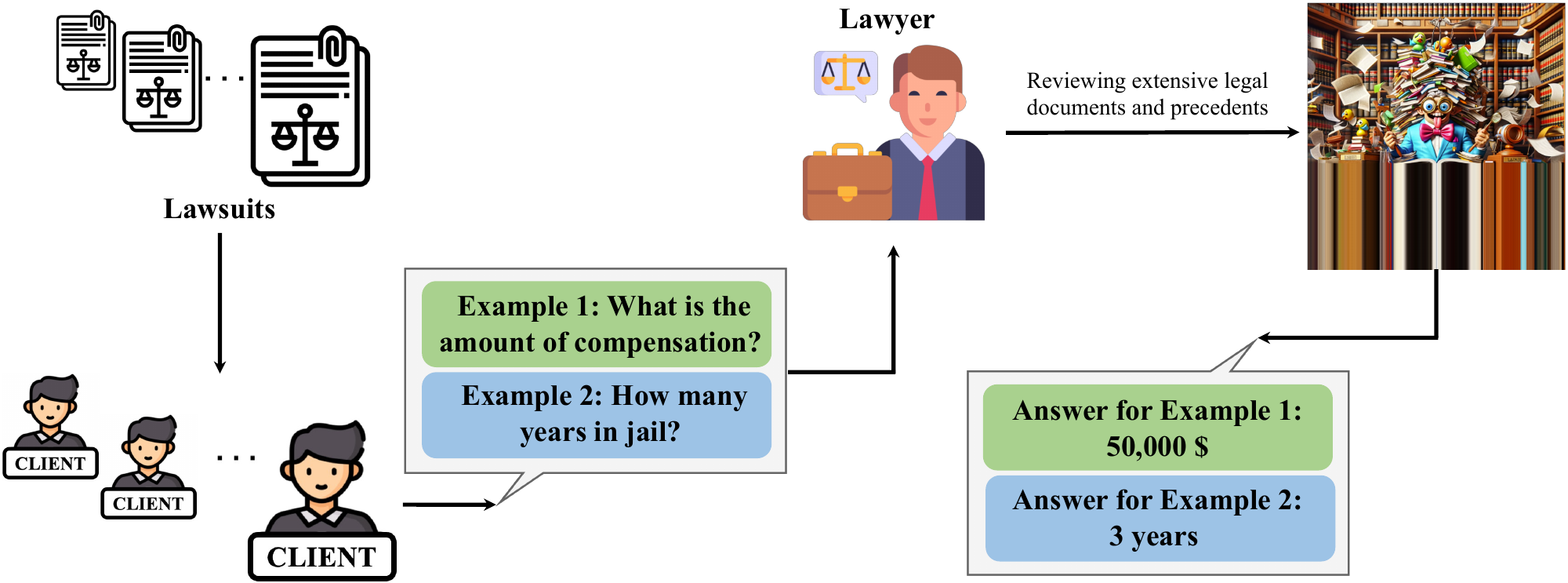}
\end{center}
\vspace{-0.2cm}
   \caption{Flowchart of practical legal proceedings. When clients or litigants bring their lawsuits to a law firm, they often pose questions pertaining to their primary concerns, such as financial compensation or imprisonment duration. However, providing prompt answers to such queries is challenging, as lawyers may need considerable time to review extensive legal documentation and precedents to offer well-informed responses. Furthermore, if the lawsuit involves numerical estimations related to other fields, such as estimating asset values, additional experts from those fields, such as asset valuation experts, are often required, further prolonging the problem-solving process.
   % This diagram illustrates the sequence from client queries regarding lawsuits to the delivery of legal outcomes. Clients present cases to a lawyer, who then reviews extensive legal documentation and precedent to provide informed answers. The examples shown demonstrate typical client questions and the corresponding legal advice or verdicts, which include financial compensation and imprisonment duration
   % \textcolor{red}{1. image captioning task flowchart. 2. proposed evaluation framework. 3. examples for exlaining why BLEU/ROUGE score is not good....4. Talbe for listing existing state-of-the-art image captioning models. 5. Human evaluations. 6. 2-3 experimental tables. 7. 2-3 visualization figures.} 
   }
% \vspace{-0.4cm}
\label{fig:figure1}
\end{figure*}

The legal field spans a broad spectrum of lawsuit types, encompassing motor vehicle accidents, personal injury compensation, estate distribution, medical malpractice, and various other areas. Lawyers encounter significant challenges in delivering prompt and accurate information to their clients across these various lawsuit types, particularly when it comes to crucial aspects like potential imprisonment duration or financial repercussions in case of unsuccessful lawsuits. This difficulty arises from the necessity to navigate extensive legal documentation and precedent to provide informed responses, referring to Figure \ref{fig:figure1}. Unfortunately, this process not only prolongs legal proceedings but also introduces inconsistencies and uncertainties in legal outcomes \cite{zhong2020does}. 
Using the U.S. as an example, compounding this issue is the fact that, according to the U.S. Financial Education Foundation (USFEF) \cite{USFEF}, over $40$ million lawsuits are filed annually in the United States. However, the total number of registered lawyers in the United States from $2007$ to $2022$ was only approximately $1.34$ million. This highlights a severe scarcity of legal experts or practitioners. Consequently, there is an urgent need to enhance the effectiveness and efficiency of traditional legal proceedings and workflows.

In recent years, spurred by rapid advancements in deep learning, researchers have increasingly turned to the application of deep learning techniques within the realm of legal Artificial Intelligence (LegalAI) \cite{zhong2020does}. This trend has led to the proposal of several new LegalAI datasets \cite{kano2019coliee,xiao2018cail2018,duan2019cjrc,chalkidis2019large,chalkidis2019deep}, which serve as pivotal benchmarks for research in this burgeoning field. Leveraging these datasets, researchers have embarked on exploring Natural Language Processing (NLP)-based solutions for a range of LegalAI tasks, including legal judgment prediction \cite{aletras2016predicting,luo2017learning,zhong2018legal,chen2019charge}, legal entity recognition and classification \cite{cardellino2017legal,angelidis2018named}, court view generation \cite{ye2018interpretable}, legal summarization \cite{hachey2006extractive,bhattacharya2019comparative}, and legal question answering \cite{monroy2009nlp,taniguchi2017legal,kim2017two}. 

However, existing methodologies in LegalAI frequently encounter challenges when attempting to accurately compute specific numerical values for legal-related estimations, such as compensations or prison durations, based on available data. This difficulty probably stems from the predominant emphasis of current legal datasets on traditional NLP text-based tasks like summarization, judgment prediction, and text-based legal question answering. Nevertheless, stakeholders in practical scenarios often demand precise numerical estimates, presenting substantial hurdles for legal professionals. Bridging this gap is crucial given the urgent demand for precision-focused applications within the realm of LegalAI.

Recent advancements in Large Language Models (LLMs) offer promising solutions to this challenge. LLMs possess the inherent capability to perform mathematical reasoning tasks with proper prompt guidance, enabling them to generate accurate numerical estimates \cite{kojima2022large,imani2023mathprompter,wei2022chain}. 
Leveraging this potential, this work proposes a novel approach that integrates LLM-based methodologies with a specially designed prompt approach. This method aims to bridge the gap between traditional legal tasks and the evolving demands of precision-oriented LegalAI applications.
To validate the efficacy of the proposed method, we introduce a curated dataset tailored to the specific needs of precision-oriented LegalAI tasks. This dataset serves as a robust benchmark for evaluating the performance of LLM-based approaches in generating accurate numerical estimates for legal scenarios. Through extensive experimentation and analysis, the results demonstrate the effectiveness of the proposed methodology in addressing the precision requirements of LegalAI applications.

\vspace{+3pt} 
\noindent The main contributions of this work are summarized as follows:
\vspace{-3pt}
\begin{itemize}
    \item \textbf{Enhanced Legal Proceedings with LLMs:} We introduce an innovative LLM-based method designed to improve the effectiveness and efficiency of traditional legal proceedings and workflows. By leveraging advanced language models, this approach also addresses the scarcity of legal experts or practitioners.

    \item \textbf{Prompt Design for Numerical Estimation Challenges:} We propose a novel approach involving specially designed prompts with in-context learning to address numerical estimation challenges within the legal domain using LLM-based models. This includes tasks such as asset valuation and imprisonment duration estimation, contributing to more informed decision-making processes in legal contexts.

    \item \textbf{Real-world Dataset of House Prices:} Introducing a real-world dataset of house prices, we provide a valuable resource for validating LLM-based methods in the domain of financial or numerical estimation within the legal field. Extensive experiments are conducted using this dataset to evaluate the effectiveness of the proposed LLM-based method.

\end{itemize}

\section{Related Work}
In this section, we explore the significant developments across three pivotal areas in the integration of AI within the legal domain. Starting with the early applications of conventional AI to tackle legal challenges, we then transition to the critical role of LegalAI datasets in refining and enhancing AI's capabilities for legal applications. Finally, we delve into the advancements brought forth by LLMs, which have reshaped the landscape of NLP and AI, offering novel methodologies and approaches in the legal field.

% \textcolor{red}{``https://github.com/thunlp/LegalPapers?tab=readme-ov-file'' talking about all the tasks metnioned in this github.}
\subsection{Conventional AI in Legal Contexts}
The intersection of AI with the legal domain has evolved significantly, transitioning from rule-based expert systems to sophisticated deep learning models. Early endeavors in LegalAI sought to replicate human legal reasoning through systems like TAXMAN and HYPO, which employed rule-based logic to navigate legal principles \cite{gardner1987artificial,rissland1987hypo}. While these systems laid the groundwork for AI's application in legal contexts, they were limited by the scope of their hardcoded knowledge and lacked the ability to generalize beyond their specific programming.

\noindent\textbf{Deep Learning and Legal Judgment Prediction (LJP):} The exploration of LJP using deep learning technology marks a significant milestone in LegalAI. The authors of \cite{hu2018few,wang2019using,liu2019legal} have pioneered this domain by employing neural networks to analyze legal documents and predict outcomes. These advancements not only showcase the potential of novel models in enhancing performance but also underscore the importance of leveraging deep learning for more accurate legal analytics.

\noindent\textbf{Innovations in Model Architectures:} The drive to improve LJP performance has led to the adoption of more sophisticated neural network architectures. For instance, the authors \cite{chen2019charge} introduced gating mechanisms to refine the prediction of penalties, while the authors of \cite{pan2019charge} developed multi-scale attention models to address the complexities of cases with multiple defendants. These innovations highlight the continuous search for model architectures that can better capture the nuances of legal reasoning.

\noindent\textbf{Utilization of Legal Knowledge:} A notable direction in recent LegalAI research is the integration of legal knowledge into AI models. In \cite{luo2017learning}, the use of attention mechanisms between facts and law articles exemplifies how AI can leverage legal knowledge to improve charge predictions. Similarly, the topological graph approach utilizes the relationships between different LJP tasks, demonstrating the value of structured legal knowledge in enhancing model performance \cite{zhong2018legal}.

\noindent\textbf{Advancements in Legal Entity Recognition and Other NLP Applications:} Beyond LJP, significant progress has been made in legal entity recognition, classification, and other NLP-based tasks. \cite{cardellino2017legal,angelidis2018named} in developing models for legal entity recognition and classification have paved the way for more sophisticated document analysis techniques. Additionally, innovations in court view generation and legal summarization by \cite{ye2018interpretable,hachey2006extractive} have opened new avenues for automating the synthesis and summarization of legal texts.

\noindent\textbf{Legal Question Answering:} The domain of legal question answering has also benefitted from AI advancements, with researchers developing models that provide precise responses to complex legal inquiries. The contributions by \cite{monroy2009nlp,taniguchi2017legal,kim2017two} in this space further bridge the gap between legal knowledge and AI capabilities, offering promising tools for legal practitioners and the public.

The shift from conventional AI to deep learning within the legal sector signals a paradigm transformation, enhancing the efficiency, accuracy, and accessibility of legal analyses and democratizing legal expertise. However, this advancement brings forth challenges such as the opaque nature of deep learning models \cite{huang2019novel_1,huang2017vqabq,huang2017robustness,huang2017robustnessMS,huang2023improving,huang2019assessing,huang2021deepopht,huang2022non,huang2021contextualized,huang2021deep,huang2021longer,wu2023expert,huck2018auto,liu2018synthesizing,yang2018novel,di2021dawn,huang2020query,huang2021gpt2mvs,huang2022causal,huang2023causalainer,huang2023query,huang2023conditional,hu2019silco,wang2024ada,zhu2024enhancing,huang2024multi,huang2024novel,zhang2024comparative,zhang2024beyond,zhang2024towards,zhang2024qfmts}, which often operate as ``black boxes'', making it difficult to discern the logic behind their decisions. This opacity poses a significant issue in the legal domain where the rationale for decisions is paramount. Additionally, deep learning models are prone to reflecting the biases present in their training data, raising ethical concerns about fairness and impartiality in AI-generated legal outcomes. Moreover, these models sometimes struggle with generalizing to new, unseen scenarios, limiting their applicability to the diverse and unique nature of legal cases.

Our approach aims to mitigate these drawbacks by harnessing the power of LLMs augmented with in-context learning and specifically designed prompts to improve interpretability, reduce bias, and enhance generalization capabilities. We focus on making the reasoning processes of LLMs more transparent, enabling legal professionals to understand and trust AI-generated insights. By implementing comprehensive bias audits and adjustments, our methodology strives to ensure ethical standards and fairness in legal AI applications. Additionally, we incorporate in-context learning to endow models with adaptability to diverse legal scenarios, bridging the divide between traditional legal practices and modern AI capabilities. This synergistic integration of human expertise and AI promises to revolutionize legal practices, making them more efficient, equitable, and adaptable to the evolving demands of the legal domain.

\vspace{-0.3cm}

% \textcolor{red}{``https://github.com/thunlp/LegalPapers?tab=readme-ov-file''  this is for the all related work.}
\subsection{LegalAI Datasets}
The evolution and impact of LegalAI have been markedly accelerated by the advent and meticulous development of various legal datasets. These datasets, each meticulously curated to reflect the intricacies of legal processes and knowledge, have been instrumental in propelling forward the application of AI in the realm of law. They serve as a critical bridge, connecting traditional legal expertise with the computational efficiency and scalability of modern AI technologies.

Among these, the JEC-QA dataset introduced by \cite{zhong2020jec} emphasizes the growing necessity for specialized datasets geared towards facilitating question-answering mechanisms within legal contexts. This need is echoed in contributions such as CAIL2019-SCM by \cite{xiao2019cail2019,duan2019cjrc}, which focus on aspects like case similarity and judicial reading comprehension, thereby underscoring the imperative to thoroughly understand and interpret judicial documents.
The contributions of \cite{chalkidis2019large,chalkidis2019neural} further expand LegalAI's research scope, introducing datasets for multi-label text classification on EU legislation and for neural legal judgment prediction in English. These efforts significantly enhance the textual analysis capabilities of AI models while broadening their application across different legal systems.
In parallel, the work \cite{manor2019plain} on summarizing contracts into plain English addresses the critical need for making legal documents more comprehensible and accessible, reinforcing AI's role in demystifying legal language. The dataset CAIL2018 by \cite{xiao2018cail2018} further exemplifies the AI's potential in automating and predicting legal outcomes, offering a robust platform for judgment prediction.
Furthermore, the LENER-BR dataset by \cite{luz2018lener}, aimed at named entity recognition in Brazilian legal texts, highlights the importance of specialized AI models for legal document analysis and information extraction. Similarly, the COLIEE dataset by \cite{kano2019coliee} and the legal case law search test collection by \cite{locke2018test} provide vital resources for evaluating legal information extraction, entailment, and the performance of legal search engines, underscoring the comprehensive nature of LegalAI studies.
Moreover, the inclusion of datasets like JURISDIC by \cite{demenko2008jurisdic} for legal dictation in Polish, the LKIF Core ontology by \cite{hoekstra2007lkif} for basic legal concepts, and the HOLJ corpus by \cite{grover2004holj} for legal text summarization, each adds a unique dimension to the LegalAI research landscape. These contributions not only aid in the development of AI models tailored for legal text processing but also highlight the interdisciplinary synergy among linguistic, legal, and computational domains.
The integration of deep learning methodologies, particularly for achieving semantic-level similarity calculations as seen in the work of \cite{tran2019building}, represents a significant stride towards refining legal information retrieval (LegalIR). The pursuit of improved embedding techniques and document-level analyses further exemplifies the ongoing efforts to enhance the precision and efficiency of legal document processing \cite{landthaler2016extending,sugathadasa2019legal}. 

This comprehensive overview underscores the significant shift towards incorporating sophisticated AI tools in legal research, emphasizing the indispensable role of meticulously designed datasets in fostering innovation and enhancing the understanding of legal reasoning and decision-making through AI. The careful development and curation of these datasets have greatly contributed to advancements in LegalIR and AI-driven legal applications.
However, existing datasets are predominantly geared towards NLP text-based tasks and lack suitability for evaluating the numerical estimation proficiency of LLM models within the legal domain. Recognizing this limitation, there is a clear need for the introduction of a real-world dataset meticulously designed to gauge LLM's efficacy in numerical estimation tasks pertinent to legal contexts, including but not limited to house value and imprisonment duration estimation. This need underscores the motivation behind our proposal for such a dataset.

\subsection{Large Language Models}
The integration of LLMs into the realm of NLP and AI marks a significant milestone in the advancement of technology. Models like GPT \cite{brown2020language} and BERT \cite{devlin2018bert} have been at the forefront of this evolution, showcasing an unprecedented ability to understand and generate human language across a variety of tasks such as translation, summarization, and question-answering \cite{brown2020language,devlin2018bert}. The significance of these developments extends beyond the enhancement of AI's linguistic capabilities, as they lay the groundwork for broader applications in multiple domains.

The scaling of LLMs, as highlighted by the works of \cite{vaswani2017attention,raffel2020exploring}, has propelled NLP to new heights, enabling machines to perform tasks with a level of sophistication that was previously unattainable. A key breakthrough in this journey has been the discovery of LLMs' inherent ability for zero-shot and few-shot learning, which has revolutionized the way these models are applied to solve tasks with minimal initial guidance \cite{liu2021makes}. This led to the emergence of ``prompting'' techniques, fundamentally altering how models are conditioned to perform specific tasks, thereby enhancing their utility across a range of applications \cite{schick2020s,reynolds2021prompt}.
Despite their successes, traditional prompting methods faced limitations in complex, multi-step reasoning tasks, leading to the development of the ``chain-of-thought (CoT)'' prompting technique. This technique, inspired by human problem-solving strategies, has significantly improved LLMs' performance in tasks requiring intricate reasoning, demonstrating the versatility and adaptability of these models in tackling more sophisticated challenges \cite{wei2022chain, wang2022self}.
Expanding the application of LLMs into mathematical reasoning, techniques like Zero-shot-CoT and MathPrompter have showcased the potential of LLMs in solving arithmetic problems and conducting mathematical analysis by generating algebraic expressions or Python functions. This not only underscores LLMs' ability in arithmetic tasks but also highlights their reliability through the validation of intermediate steps, as evidenced by their performance on the MultiArith dataset \cite{kojima2022large, imani2023mathprompter}.

% The advancements in LLM technologies signify a broader shift towards more dynamic and intelligent AI applications, extending their impact from linguistic to non-linguistic domains, including the legal sector. Our initiative, aimed at enhancing decision-making and operational efficiency in the legal domain through LLMs, seeks to leverage these advancements to address the unique challenges faced by legal professionals. By incorporating LLMs' mathematical reasoning capabilities with specially designed prompts, we aim to improve the precision and efficiency of legal workflows, bridging the gap between traditional legal practices and modern technological innovations. The introduced dataset tailored for precision-oriented LegalAI tasks further validates our methodology, confirming the potential of LLMs to revolutionize legal processes and contribute to a more streamlined, accessible, and equitable legal system.

The progress in LLM technologies marks a significant shift towards more versatile and intelligent AI applications, expanding their influence from linguistic to non-linguistic domains, particularly in the legal sector. Our initiative, aimed at improving decision-making and operational efficiency within the legal domain through LLMs, harnesses these advancements to tackle the distinct challenges encountered by legal professionals. By integrating LLMs' mathematical reasoning capabilities with tailored prompts, our goal is to enhance the precision and efficacy of legal workflows, thereby bridging the traditional methods of legal practice with contemporary technological innovations. The introduction of a curated dataset tailored for precision-focused LegalAI tasks further substantiates our approach, affirming the potential of LLMs to revolutionize legal processes and contribute to a more streamlined, accessible, and equitable legal system.

% \textcolor{red}{The proposed LLM-based approach contributes to a more streamlined, accessible, and equitable legal system. Moreover, by enhancing the interpretability and transparency of AI applications in legal contexts, this method plays a crucial role in addressing the inherent opacity associated with the current applications of deep learning-based models.} 

%%%%%%%%%%%%%%%%%%%%%%%%%%%%%%%%%%%%%%%%%%%%%%%%%
%%%%%%%%%%%%      Methodology       %%%%%%%%%%%%%
%%%%%%%%%%%%%%%%%%%%%%%%%%%%%%%%%%%%%%%%%%%%%%%%%

% \vspace{-0.1cm}
\section{Methodology}
The proposed method encompasses several essential components, including in-context learning and specially designed prompts. Each of these components will be  introduced in the subsequent subsections. Additionally, to validate the effectiveness of the proposed method for precision-oriented LegalAI tasks, we propose a dataset focused on asset value estimation within the legal domain.

% \textcolor{red}{maybe draw a flowchart figure for the method? or just skip it? think about it later. check other in-context learning papers about their method flowchart.}

\subsection{In-context Learning}
GPT-3 \cite{brown2020language} and other LLMs have exhibited remarkable proficiency in few-shot predictions without requiring fine-tuning. This entails providing the model with a task description in natural language along with a small number of examples. The effectiveness of this learning capability hinges on scaling model size, data, and computing resources. \cite{rae2021scaling,smith2022using,chowdhery2023palm,du2022glam} have proposed various training methodologies for different types of LLMs. These models exhibit a notable capacity to leverage few-shot prompts for accomplishing unseen tasks without the necessity of fine-tuning, a capability that emerges prominently in LLMs compared to their smaller counterparts.
While LLMs \cite{brown2020language,chowdhery2023palm} have showcased remarkable success across various NLP tasks, their aptitude for reasoning has often been viewed as limited. It's important to note that merely scaling up the model size does not necessarily enhance this capability.

Recent advancements have introduced the concept of ``chain of thoughts'' prompting, also known as in-context learning, as a potent technique to augment the reasoning prowess of LLMs when processing text \cite{wei2022chain}. This method entails presenting the model with multiple instances of reasoning chains, enabling LLMs to learn and apply the underlying template to solve intricate, unseen tasks effectively. Notably, the ``chain of thoughts'' approach bears resemblance to the strategy employed in the Visual Question Answering task, where addressing basic questions aids in tackling complex queries \cite{huang2019novel}.
These recent findings underscore the robust reasoning capabilities of LLMs. Nevertheless, the current research landscape in LegalAI predominantly concentrates on text-based tasks such as legal question answering and legal summarization. In contrast, this paper aims to investigate the capacity of LLMs, coupled with in-context learning, to tackle precision-oriented tasks through meticulously designed prompts. The details of these crafted prompts are elucidated in the subsequent subsection.

\begin{figure}[t!]
    \centering
    \begin{subfigure}[b]{0.5\textwidth}
        \centering
        \includegraphics[width=\textwidth]{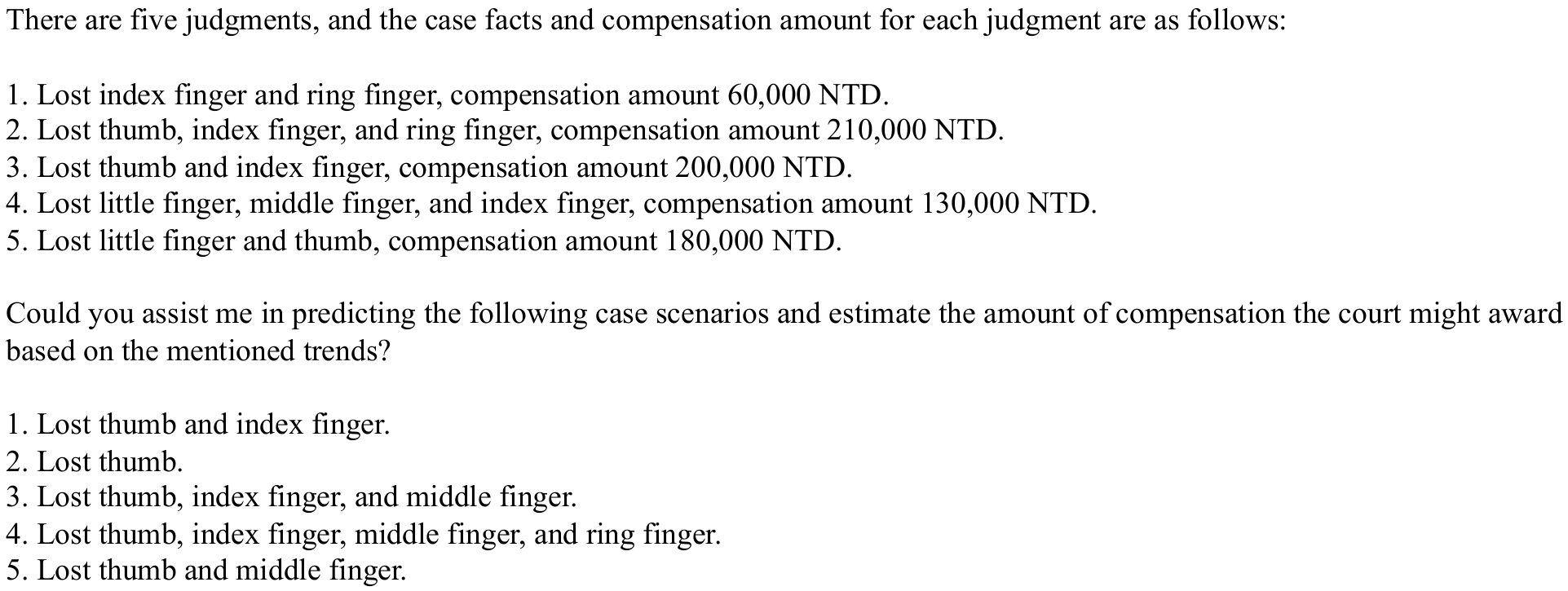}
        \caption{Example prompt design for predicting compensation amount in the legal domain.}
        \label{fig:sub11}
    \end{subfigure}
    \hfill
    \begin{subfigure}[b]{0.5\textwidth}
        \centering
        \includegraphics[width=\textwidth]{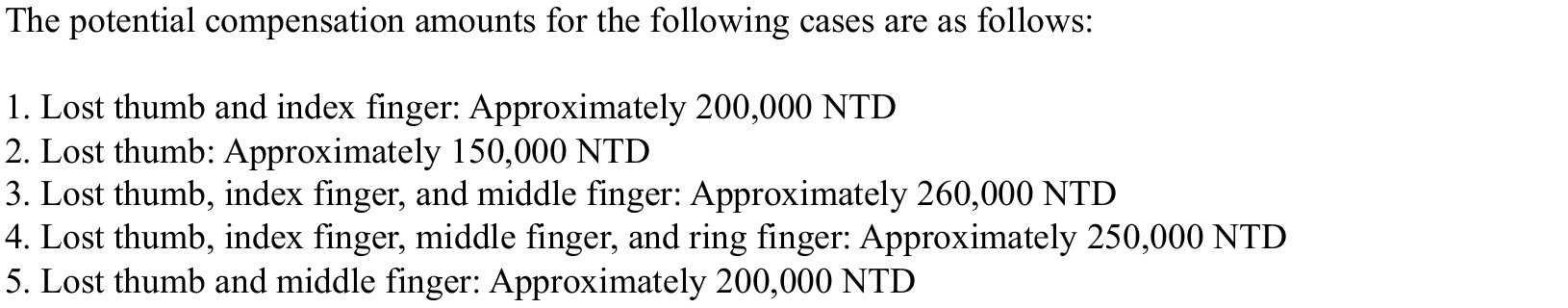}
        \caption{The output is generated by GPT-4 in response to (a).}
        \label{fig:sub22}
    \end{subfigure}
    \vspace{-0.6cm}
    \caption{Design of an example prompt for compensation amount estimation, as depicted in (a). Additionally, the corresponding response generated by GPT-4 is included to ensure the completeness of this question-answering conversion, referenced in (b). The ground truth answers for each judgment in the example are $200,000$ NTD, $150,000$ NTD, $250,000$ NTD, $260,000$ NTD, and $200,000$ NTD, respectively.
}
    \label{fig:figure4}
    % \vspace{-0.6cm}
\end{figure}

\subsection{Prompts Design}
In this study, we advocate for the utilization of in-context learning coupled with specially formulated prompts to address precision-oriented tasks within the legal domain. Our prompts are designed with specific examples that encompass various attributes, including house specifications like area, location, and type, as well as details related to scam amounts or injury locations/parts. Each example is accompanied by corresponding answers such as house prices, compensation values, or durations of imprisonment. These examples serve as training demonstrations for LLMs to predict outputs for unseen test examples.

\begin{figure}[t!]
    \centering
    \begin{subfigure}[b]{0.5\textwidth}
        \centering
        \includegraphics[width=\textwidth]{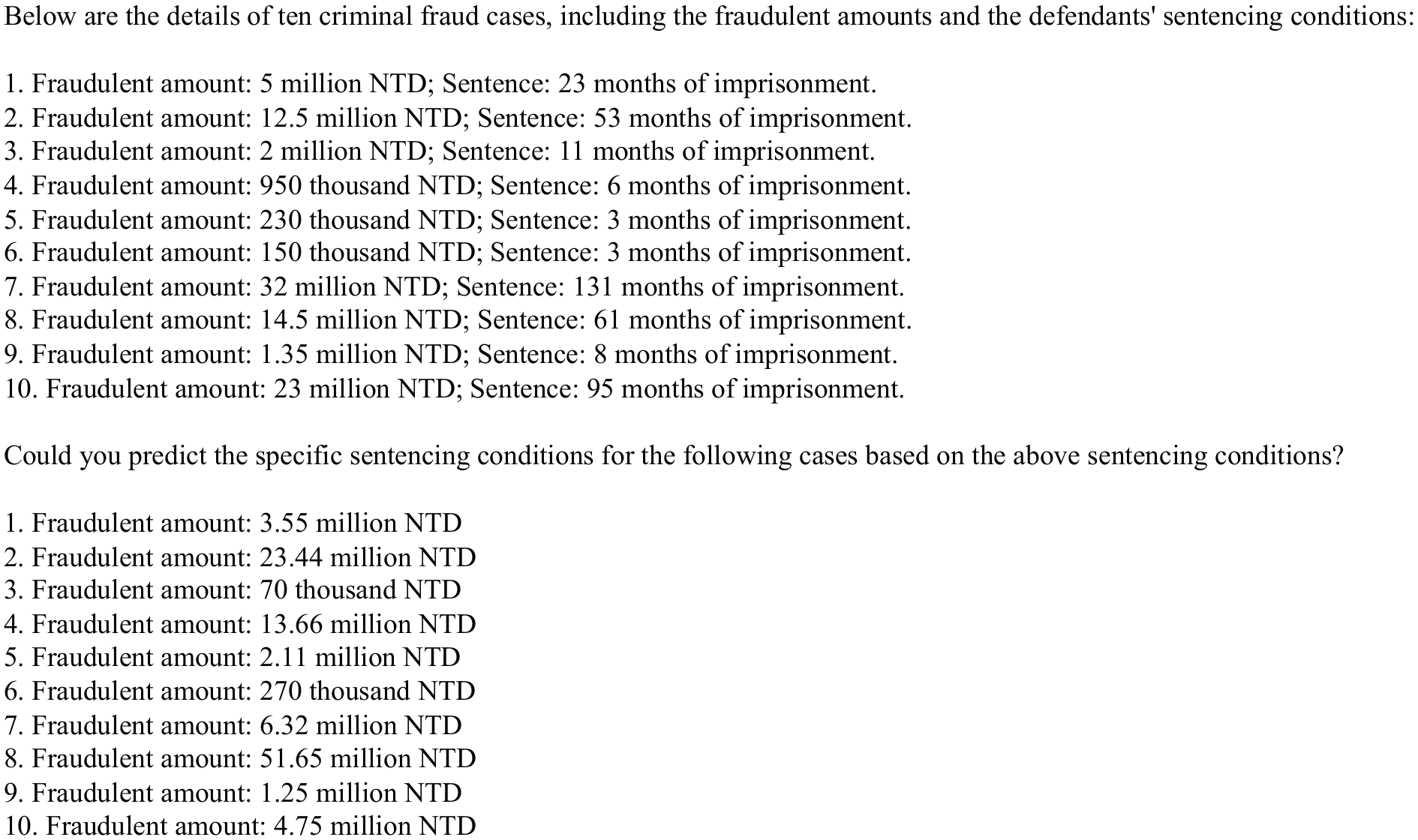}
        \caption{Example prompt design for predicting imprisonment duration in the legal domain.}
        \label{fig:sub1}
    \end{subfigure}
    \hfill
    \begin{subfigure}[b]{0.5\textwidth}
        \centering
        \includegraphics[width=\textwidth]{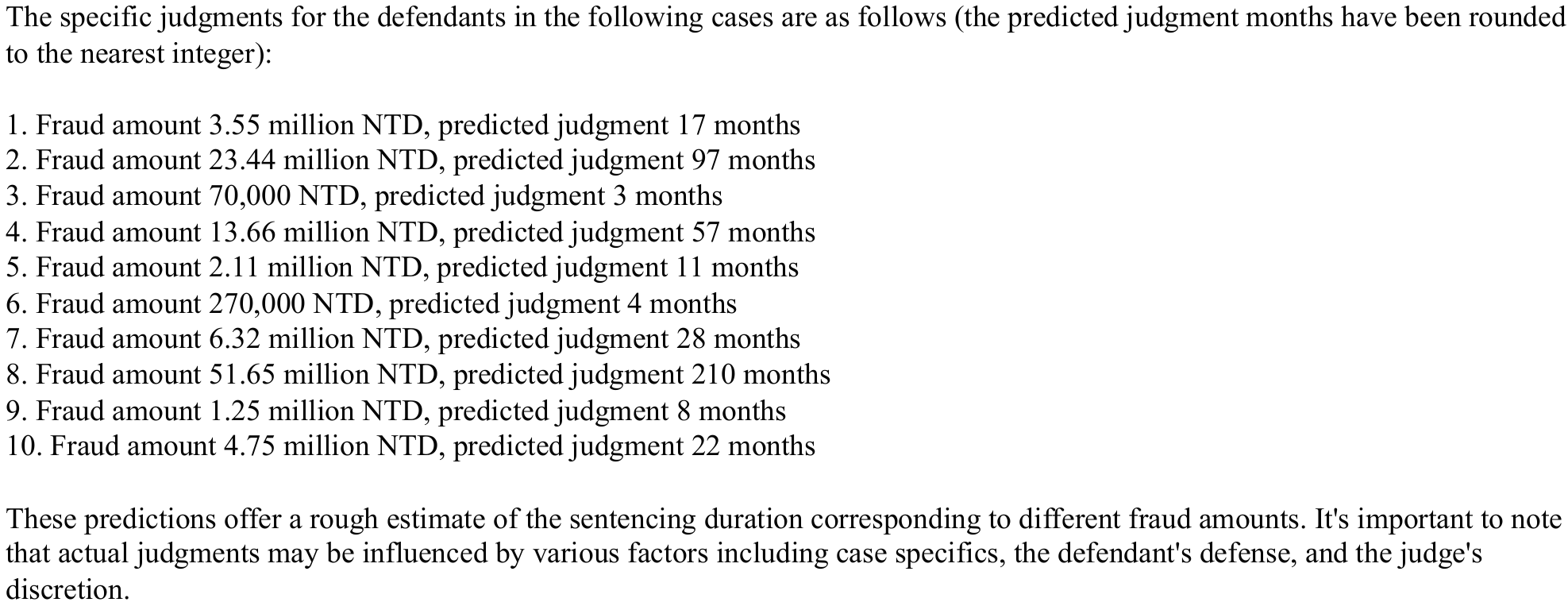}
        \caption{The output is generated by GPT-4 in response to (a).}
        \label{fig:sub2}
    \end{subfigure}
    \vspace{-0.6cm}
    \caption{Creation of an example prompt for imprisonment duration estimation, illustrated in (a). Furthermore, the corresponding response generated by GPT-4 is provided to ensure the comprehensiveness of this question-answering conversion, as indicated in (b). The ground truth answers for each judgment in the example are as follows: 17.2 months, 96.76 months, 3.28 months, 57.64 months, 11.44 months, 4.08 months, 28.28 months, 209.6 months, 8 months, and 22 months, respectively.}
    \label{fig:figure4-1}
    % \vspace{-0.6cm}
\end{figure}

Consider the prompt format illustrated in Figure \ref{fig:figure4} as a case in point. The first step in designing prompts entails extracting pivotal information from each judgment, ensuring a uniform textual description format encompassing information on injuries and compensation amounts. Following this, the next step involves framing a question that pertains to forecasting trends based on the provided cases. The proposed prompt format is devised to aid the LLM in identifying patterns within the presented cases. By leveraging these discerned patterns, the LLM is empowered to make predictions for the given test cases.
If the variable of interest transitions from compensation to imprisonment duration, the procedure remains consistent, as depicted in Figure \ref{fig:figure4-1}.

% \begin{figure}[ht!]
% \begin{center}
% \includegraphics[width=1.0\linewidth]{finger_example.pdf}
% \end{center}
% \vspace{-0.4cm}
%    \caption{Example prompt design for estimating compensation amount in the legal domain. \textcolor{red}{add the prediction results. GPT-4 and Gemini?}}
% \vspace{-0.4cm}
% \label{fig:figure4}
% \end{figure}

% \begin{figure}[ht!]
% \begin{center}
% \includegraphics[width=1.0\linewidth]{finger_example_prediction.pdf}
% \end{center}
% \vspace{-0.4cm}
%    \caption{The output is generated by GPT-4 in response to Figure \ref{fig:figure4}.}
% \vspace{-0.4cm}
% \label{fig:figure4-ans}
% \end{figure}

\subsection{Proposed Precision-Oriented LegalAI Dataset}
\begin{figure}[t!]
\begin{center}
\includegraphics[width=1.0\linewidth]{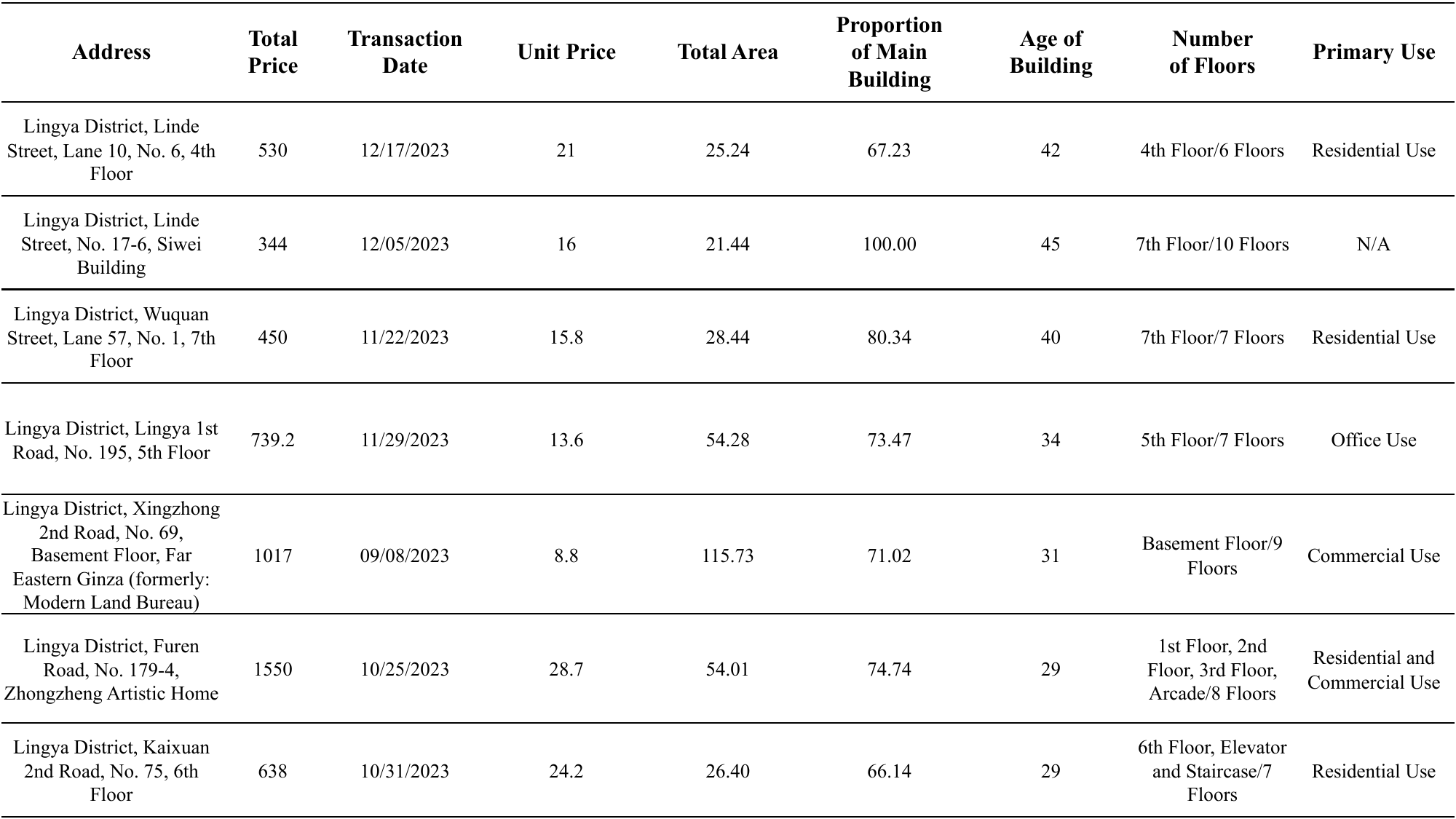}
\end{center}
\vspace{-0.2cm}
   \caption{Randomly selected examples from the proposed dataset. Each row represents a data sample with nine different properties.}
% \vspace{-0.6cm}
\label{fig:figure3}
\end{figure}

The valuation of properties, encompassing houses, lands, and various assets, is a prevalent practice in the legal domain, particularly in cases involving legacy or property division disputes. 
However, this process is typically time-consuming and resource-intensive due to the participation of numerous experts in the estimation phase. The complexity and duration of the overall process are further compounded by the involvement of experts specializing in property valuation.  Consequently, there is an urgent need for an effective method to streamline and enhance the entire valuation process. Using house value estimation as an illustrative case, this study introduces a dataset specifically tailored to the common and crucial task of estimating house values. 
This dataset serves to assess the efficacy of the proposed method in addressing the precision requirements inherent in such tasks within the legal domain.
The presented dataset comprises $58$ data samples, with $45$ samples allocated for in-context learning and the remaining $13$ utilized for testing purposes. Each house data sample within the proposed dataset includes essential information and attributes such as address, total price, transaction date, unit price, total area, proportion of the main building, age of the building, number of floors, and primary use.
To facilitate comprehension of the dataset, several examples from it are illustrated in Figure \ref{fig:figure3}.

% \vspace{-0.1cm}
\section{Experiments and Analysis}
In this section, our objective is to assess the efficacy of the proposed method tailored for numerical estimation tasks within the legal domain. To accomplish this, we will conduct a validation of our method using the real-world dataset we have introduced for housing price estimation within the legal domain. The specifics of this dataset are outlined in the Methodology section.

\subsection{Experimental Settings}

Acquiring authentic and up-to-date data related to houses within legal domain poses challenges, resulting in a relatively modest size for the proposed dataset. Despite its smaller scale, this dataset proves sufficient for the purpose of in-context learning \cite{wei2022chain}. In the in-context learning phase, various attributes such as address, total price, transaction date, unit price, total area, proportion of the main building, age of the building, number of floors, and primary use are provided. In the testing phase, only the attributes of address, total area, proportion of the main building, age of the building, number of floors, and primary use are supplied. The model is tasked with predicting the total price during the testing phase. Once the total price is predicted, the unit price can be determined based on the available information of the total area and the predicted total price.

We conduct experiments using state-of-the-art LLMs, including OpenAI GPT-3.5, OpenAI GPT-4, Claude AI, and Google Bard with Gemini. The error rate (ER) and mean absolute percentage error (MAPE) metrics are employed to quantify the model's performance. MAPE is a measure used to assess the accuracy of a predictive model, particularly in forecasting tasks. It calculates the average absolute percentage difference between the predicted values and the actual values. The absolute values of errors are taken to prevent positive and negative errors from canceling each other out. MAPE is expressed as a percentage, and lower values indicate better predictive accuracy. The formula for calculating MAPE is provided in Equation (\ref{eqn:mape}).
\begin{equation}
\text{ER} = \frac{G_t - P_t}{G_t} ; \text{MAPE} = \frac{1}{n} \sum_{t=1}^{n} \left| \text{ER} \right| \times 100\%,
\label{eqn:mape}
\end{equation}
where $G_t $ is the ground truth value at time $t$, $P_t$ is the predicted value at time $t$, and $n$ is the total number of observations.

% \textcolor{red}{Mean absolute percentage error measures the average magnitude of error produced by a model, or how far off predictions are on average. A MAPE value of 20\% means that the average absolute percentage difference between the predictions and the actuals is 20\%. In other words, the model’s predictions are, on average, off by 20\% from the real values. A lower MAPE value indicates a more accurate prediction – an MAPE of 0\% means the prediction is the same as the actual, while a higher MAPE value indicates a less accurate prediction. 
% To calculate the mean absolute percentage error, we first calculate the absolute value of all the residuals. We take the absolute value of the errors because we do not want positive and negative errors to cancel each other out. If our model overshoots one data point by +10 and undershoots another by -10, these errors would cancel each other out because (-10) + 10 = 0. But by taking the absolute value of the errors we avoid this problem, because $|10| + |-10| = 20.$}

\subsection{Evaluation and Analysis}

\noindent\textbf{OpenAI GPT-3.5.}

GPT-3.5 stands as a sophisticated iteration of OpenAI's GPT language model, following the achievements of its predecessor, GPT-3. This model is engineered to produce text that closely resembles human language in response to given inputs. Employing a transformer architecture, GPT-3.5 undergoes pre-training on an extensive corpus of internet text, enhancing its capacity to comprehend and generate coherent responses across diverse domains. Notably, it exhibits refined capabilities in contextual understanding, nuanced response generation, and adept handling of intricate language tasks.
In this study, we harness GPT-3.5 for in-context learning, utilizing a meticulously crafted prompt detailed in the Methodology section. The efficacy of this approach is then evaluated using our proposed precision-oriented dataset. As shown in Table \ref{table:table1}, the MAPE value is $40.75\%$. This outcome suggests that there is room for improvement in GPT-3.5's numerical estimation capabilities.

% ``mean error = - 10.88\%''
\begin{table}[t!]
    \centering
    \caption{Results for OpenAI GPT-3.5 show a MAPE value of $40.75\%$. The ``Difference'' is defined as ``Ground Truth (GT) $-$ Prediction (Pred)''.}
    \vspace{-0.4cm}
    \begin{tabular}{ccccc}
        \toprule
        \textbf{Test Case} & \textbf{Pred} & \textbf{GT} & \textbf{Difference} & \textbf{ER (\%)} \\
        \midrule
        Case 1 & 480 & 920 & 440 & 47.8 \\
        Case 2 & 1,002 & 650 & -352 & -54.2 \\
        Case 3 & 1,118 & 700 & -418 & -59.7 \\
        Case 4 & 297 & 400 & 103 & 25.8 \\
        Case 5 & 542.7 & 450 & -92.7 & -20.6 \\
        Case 6 & 285.97 & 315 & 29.03 & 9.2 \\
        Case 7 & 455.24 & 710 & 254.76 & 35.9 \\
        Case 8 & 1,008.10 & 750 & -258.1 & -34.4 \\
        Case 9 & 421.99 & 730 & 308.01 & 42.2 \\
        Case 10 & 830.92 & 790 & -40.92 & -5.2 \\
        Case 11 & 701.76 & 440 & -261.76 & -59.5 \\
        Case 12 & 626.24 & 310 & -316.24 & -102 \\
        Case 13 & 546.72 & 820 & 273.28 & 33.3 \\
        \bottomrule
    \end{tabular}
\label{table:table1}
\vspace{-0.5cm}
\end{table}

\noindent\textbf{OpenAI GPT-4.}

GPT-4 represents the latest advancement in OpenAI's series of LLMs, incorporating multimodal capabilities. This iteration, following in the footsteps of its predecessors, employs a transformer-based architecture and a training paradigm that combines public data with third-party licensed data to predict the next token. Subsequently, the model undergoes fine-tuning through reinforcement learning, incorporating feedback from both human and AI sources to ensure alignment with human language nuances and policy compliance. The implementation of GPT-4 in ChatGPT marks an evolution from the previous version based on GPT-3.5, albeit with some persisting challenges. Notably, GPT-4 introduces vision capabilities (GPT-4V), enabling the model to process image inputs within the ChatGPT framework.
To assess GPT-4's performance on our proposed dataset, we conduct experiments using the same prompt design as with GPT-3.5. As indicated in Table \ref{table:table2}, the MAPE value is $15.71\%$, significantly lower than that of GPT-3.5. These results demonstrate a notable improvement in GPT-4's numerical estimation capabilities compared to GPT-3.5.

% ``mean error = 5.52\%''
\begin{table}[t!]
    \centering
    \caption{Results for GPT-4 show a MAPE value of $15.71\%$.}
    \vspace{-0.4cm}
    \begin{tabular}{ccccc}
        \toprule
        \textbf{Test Case} & \textbf{Pred} & \textbf{GT} & \textbf{Difference} & \textbf{ER (\%)} \\
        \midrule
        Case 1 & 863.33 & 920 & 56.67 & 6.2 \\
        Case 2 & 597.87 & 650 & 52.13 & 8.0 \\
        Case 3 & 799.36 & 700 & -99.36 & -14.2 \\
        Case 4 & 358.81 & 400 & 41.19 & 10.3 \\
        Case 5 & 400.96 & 450 & 49.04 & 10.9 \\
        Case 6 & 390.42 & 315 & -75.42 & -23.9 \\
        Case 7 & 486.32 & 710 & 223.68 & 31.5 \\
        Case 8 & 818.69 & 750 & -68.69 & -9.2 \\
        Case 9 & 703.29 & 730 & 26.71 & 3.7 \\
        Case 10 & 839.22 & 790 & -49.22 & -6.2 \\
        Case 11 & 391.67 & 440 & 48.33 & 11 \\
        Case 12 & 349.52 & 310 & -39.52 & -12.7 \\
        Case 13 & 357.91 & 820 & 462.09 & 56.4 \\
        \bottomrule
    \end{tabular}
\label{table:table2}
% \vspace{-0.5cm}
\end{table}

\noindent\textbf{Claude AI.}

Claude, an AI assistant or chatbot, was developed by Anthropic and initially launched in March 2023 with the Claude 1.3 language model. However, a subsequent version, powered by the Claude 2 language model, was released in July 2023. In May 2023, Claude expanded its context window, enabling businesses to submit extensive documentation for analysis. Anthropic claims that this broader context window helps mitigate hallucination rates and enhances the model's proficiency in complex reasoning, making it a more robust and powerful solution overall.
In our experiment, our focus extends beyond the GPT-based model series to evaluate the performance of the Claude AI chatbot. To ensure a fair comparison, we maintain consistency in prompt design across all experiments. According to the results presented in Table \ref{table:table3}, the MAPE value is $29.06\%$. This outcome suggests that Claude AI's numerical estimation ability falls between that of GPT-3.5 and GPT-4.

\noindent\textbf{Google Bard with Gemini.}

Google Bard is an AI-driven chatbot tool developed by Google to emulate human conversations through NLP and machine learning. It debuted in December 2023, and in February 2024, Google expanded Gemini Pro to all supported languages within Bard, introducing text-to-image generation capabilities. Apart from enhancing Google search results, Bard can be seamlessly integrated into websites, messaging platforms, or applications to deliver lifelike, natural language responses to user inquiries. The integration of Google Gemini Pro with Bard signifies a significant advancement in user interaction. Gemini's multimodal processing abilities empower Bard to handle images, audio, and video in addition to text, fostering a more organic and engaging conversational experience. This collaborative synergy between Google Gemini and Bard underscores the evolution of conversational AI, providing users with a seamless and enriched communication platform.
Given that Google Bard with Gemini represents another state-of-the-art LLM, our experiments include a comparative analysis of model performance. Specifically, we assess the performance of GPT-3.5, GPT-4, and Claude AI in comparison to Google Bard with Gemini using our proposed dataset as a benchmark. As shown in Table \ref{table:table4} and Table \ref{table:table5}, the MAPE value of Google Bard with Gemini without internet access is $18.75\%$, while with internet access, it is $17.84\%$. These results indicate that internet access can enhance the performance of the LLM.

Comparing the results across Table \ref{table:table1}, Table \ref{table:table2}, Table \ref{table:table3}, Table \ref{table:table4}, and Table \ref{table:table5}, we conclude that GPT-4 demonstrates the best performance, with Google Bard with Gemini offering competitive performance comparable to GPT-4.

% ``mean error = 21\%'' 
\begin{table}[t!]
    \centering
    \caption{Results for Claude AI show a MAPE value of $29.06\%$.}
    \vspace{-0.4cm}
    \begin{tabular}{ccccc}
        \toprule
        \textbf{Test Case} & \textbf{Pred} & \textbf{GT} & \textbf{Difference} & \textbf{ER (\%)} \\
        \midrule
        Case 1 & 431 & 920 & 489 & 53.2 \\
        Case 2 & 681 & 650 & -31 & -4.8 \\
        Case 3 & 927 & 700 & -227 & -32.4 \\
        Case 4 & 241 & 400 & 159 & 39.8 \\
        Case 5 & 269 & 450 & 181 & 40.2 \\
        Case 6 & 306 & 315 & 9 & 2.9 \\
        Case 7 & 492 & 710 & 218 & 30.7 \\
        Case 8 & 864 & 750 & -114 & -15.2 \\
        Case 9 & 569 & 730 & 161 & 22.1 \\
        Case 10 & 730 & 790 & 60 & 7.6 \\
        Case 11 & 285 & 440 & 155 & 35.2 \\
        Case 12 & 254 & 310 & 56 & 18.1 \\
        Case 13 & 200 & 820 & 620 & 75.6 \\
        \bottomrule
    \end{tabular}
\label{table:table3}
% \vspace{-0.5cm}
\end{table}

% Given that Google Bard with Gemini represents another state-of-the-art LLM, our experiments involve a comparative analysis of model performance. Specifically, we assess the performance of the LLMs—GPT-3.5, GPT-4, and Claude AI—in relation to Google Bard with Gemini. This evaluation is conducted using our proposed dataset as a benchmark. In Table \ref{table:table4} and Table \ref{table:table5}, the MAPE value of Google Bard with Gemini without internet is $18.75\%$, and the MAPE value of Google Bard with Gemini with internet is $17.84\%$. The results show that the internet can help the LLM performs better.
% Comparing the results of Table \ref{table:table1}, Table \ref{table:table2}, Table \ref{table:table3}, Table \ref{table:table4}, and Table \ref{table:table5}, we conclude that GPT-4 has the best performance and the performance of Google Bard with Gemini is competative to GPT-4.

% \subsection{Demo how to use the proposed framework, use that hongyi image captioning blip model? and show the whole MSCOCO dataset result and other scores,... check his ipython.}
% \textcolor{red}{three results figures, 1. mscoco, 2. flickr, 3. ours.}

\section{Discussion}

% The article explores three applications that integrate law with Large Language Models (LLMs): 1. Predicting compensation amounts in civil cases, particularly finger injury cases. 2. Evaluating housing market prices for various legal purposes. 3. Forecasting the severity of criminal sentences in fraud cases.

% In the first application, LLMs are utilized to forecast compensation amounts in civil cases, where the determination often depends on specific factors like the type of damage or injuries sustained. Despite challenges in accurately estimating compensation, leveraging LLMs and historical judgment data aids lawyers in assessing potential outcomes and devising legal strategies.

% The second application focuses on determining housing market values, benefiting cases involving real estate issues and assisting in property division or compensation scenarios. This method is adaptable to determining land market values as well.

% In the third application, LLMs are employed to predict the severity of criminal sentences in fraud cases, considering factors such as the nature of the crime and the defendant's behavior post-crime. Despite potential deviations from past judgment outcomes, LLMs enhance the quality of legal advice provided by objectively predicting sentencing outcomes.

% While these applications do not replace lawyers, they serve as valuable tools for optimizing research and decision-making processes, reducing time consumption, and easing the financial burden of legal fees on litigants.

In this work, we present three applications that harness the integration of law and LLM technology: 1. Predicting compensation amounts in finger injury cases, 2. Forecasting housing market prices, and 3. Evaluating sentencing in fraud cases.

The first application involves predicting compensation amounts in civil cases, particularly in scenarios like car accidents or property damage disputes. Determining compensation often hinges on various factors, such as the type of vehicle involved in accidents or the severity of bodily injuries sustained. Despite having detailed case information, lawyers may still struggle to accurately estimate compensation, leading to challenges in decision-making regarding settlement or litigation. By leveraging LLMs and extensive historical judgment data, this approach provides a more objective insight into compensation trends. This aids lawyers in promptly assessing potential judgment outcomes and refining legal strategies. While this paper primarily examines the correlation between finger injuries and compensation amounts, this methodology can be extended to other civil case types or factors influencing judgment amounts.

The second application focuses on assessing housing market values, which proves valuable not only in cases concerning damaged real estate but also in scenarios involving property division or compensation for property encroachment. In these situations, property market values play a crucial role in determining case outcomes. Furthermore, the versatility of the proposed method extends to determining land market values, highlighting its broad applicability in legal practice.

The third application focuses on predicting the severity of criminal sentences in criminal cases, which is typically influenced by factors such as the nature and magnitude of the crime, as well as the defendant's post-crime conduct. Despite the availability of detailed case specifics, lawyers often face challenges in accurately predicting sentence severity. Leveraging LLMs to analyze past judgment data and discern the relationship between specific factors and sentence severity enables a more objective prediction of sentencing outcomes, thereby enhancing the quality of legal advice provided. While this study concentrates on predicting sentence severity based on fraud amounts, this research direction is poised to expand to encompass other factors or criminal areas.

However, trends in judgments may not remain static, and unique factors in individual cases could deviate from past judgment outcomes. These nuances still necessitate resolution by experienced lawyers. Therefore, the current applications do not aim to replace lawyers but rather serve as valuable tools for optimizing research and decision-making processes, reducing time consumption, and alleviating the financial burden of legal fees on litigants.

% ``mean error = 8.18\%'' 
\begin{table}[t!]
    \centering
    \caption{Results for Google Bard with Gemini (without internet) indicate a MAPE value of $18.75\%$.}
    \vspace{-0.4cm}
    \begin{tabular}{ccccc}
        \toprule
        \textbf{Test Case} & \textbf{Pred} & \textbf{GT} & \textbf{Difference} & \textbf{ER (\%)} \\
        \midrule
        Case 1 & 486.2 & 920 & 433.8 & 47.2 \\
        Case 2 & 714.2 & 650 & -64.2 & -9.9 \\
        Case 3 & 797.4 & 700 & -97.4 & -13.9 \\
        Case 4 & 349.2 & 400 & 50.8 & 12.7 \\
        Case 5 & 391.1 & 450 & 58.9 & 13.1 \\
        Case 6 & 383.2 & 315 & -68.2 & -21.7 \\
        Case 7 & 566.6 & 710 & 143.4 & 20.2 \\
        Case 8 & 795.2 & 750 & -45.2 & -6 \\
        Case 9 & 788.9 & 730 & -58.9 & -8.1 \\
        Case 10 & 841.7 & 790 & -51.7 & -6.5 \\
        Case 11 & 356.4 & 440 & 83.6 & 19 \\
        Case 12 & 318.2 & 310 & -8.2 & -2.6 \\
        Case 13 & 304.6 & 820 & 515.4 & 62.9 \\
        \bottomrule
    \end{tabular}
\label{table:table4}
% \vspace{-0.3cm}
\end{table}

\section{Conclusion and Future Work}
The legal field faces substantial challenges in delivering timely and accurate information across a spectrum of lawsuit types, exacerbated by the intricacies of legal documentation and the limited availability of legal experts. This highlights the urgent requirement for more efficient and effective legal procedures. While recent advancements in deep learning have prompted researchers to investigate solutions for various LegalAI tasks to enhance the efficacy of legal processes, challenges remain in accurately computing specific numerical values for legal-related estimations.
To address this gap, our study proposes an innovative approach that integrates LLMs with specially designed prompts tailored to precision-oriented LegalAI applications. This approach seeks to enhance traditional legal proceedings and workflows while meeting the demand for precise numerical estimates in practical scenarios.
The contributions of our work include introducing an LLM-based method to enhance legal proceedings, proposing a novel prompt design for addressing financial or precision-oriented challenges within the legal domain, presenting a real-world dataset of house prices for validating LLM-based methods, and conducting extensive experiments to evaluate the efficacy of the proposed approach in estimating parameters relevant to legal proceedings.
Our findings underscore the potential of LLM-based methods in meeting the precision requirements of LegalAI applications, offering promising solutions to the challenges encountered in the legal domain.
In summary, the suggested LLM-based approach aims to establish a more streamlined, accessible, and equitable legal system. Moreover, the proposed LLM-based method shows significant potential in improving the interpretability and transparency of AI applications within legal contexts, thereby playing a pivotal role in alleviating the inherent opacity associated with current applications of deep learning-based models.

Potential future research directions using LLM-based methods include exploring the correlation between specific factors and compensation judgments in diverse civil cases, such as car accidents, property damage, joint real estate division, and property misappropriation cases. Additionally, studies could investigate the correlation between specific factors and sentencing severity in various criminal cases, including embezzlement, breach of trust, property crimes, and other case-specific characteristics.

% ``mean error = -2.68\%'' 
\begin{table}[t!]
    \centering
    \caption{Results for Google Bard with Gemini (with internet) reveal a MAPE value of $17.84\%$.}
    \vspace{-0.4cm}
    \begin{tabular}{ccccc}
        \toprule
        \textbf{Test Case} & \textbf{Pred} & \textbf{GT} & \textbf{Difference} & \textbf{ER (\%)} \\
        \midrule
        Case 1 & 650 & 920 & 270 & 29.3 \\
        Case 2 & 780 & 650 & -130 & -20 \\
        Case 3 & 700 & 700 & 0 & 0 \\
        Case 4 & 420 & 400 & -20 & -5 \\
        Case 5 & 420 & 450 & 30 & 6.7 \\
        Case 6 & 450 & 315 & -135 & -42.9 \\
        Case 7 & 630 & 710 & 80 & 11.3 \\
        Case 8 & 950 & 750 & -200 & -26.7 \\
        Case 9 & 730 & 730 & 0 & 0 \\
        Case 10 & 900 & 790 & -110 & -13.9 \\
        Case 11 & 450 & 440 & -10 & -2.3 \\
        Case 12 & 380 & 310 & -70 & -22.6 \\
        Case 13 & 400 & 820 & 420 & 51.2 \\
        \bottomrule
    \end{tabular}
\label{table:table5}
% \vspace{-0.6cm}
\end{table}

% Potential future research directions using the LLM-based method in this article may include:

% 1. Studies on the correlation between specific factors and compensation judgments in diverse civil cases, such as:
%    1.1. Car accident compensation cases: Predicting amounts based on injuries and car value.
%    1.2. Property damage cases: Predicting amounts based on property characteristics.
%    1.3. Cases involving joint real estate division: Predicting market prices based on property characteristics for estimating co-owners' amounts.
%    1.4. Property misappropriation cases: Predicting compensation amounts based on property characteristics.

% 2. Studies on the correlation between specific factors and sentencing in various criminal cases, such as:
%    2.1. Predicting severity in embezzlement, breach of trust, or other property crimes based on criminal proceeds.
%    2.2. Predicting severity in property crimes or other crimes based on factors like criminal motives, methods, admission of guilt, settlement, or other case-specific characteristics.

%
% The next two lines define the bibliography style to be used, and the bibliography file.
\bibliographystyle{ACM-Reference-Format}
\bibliography{sample-base}

% 
% If your work has an appendix, this is the place to put it.
\appendix

\end{document}